\documentclass[11pt,a4paper]{article}
\usepackage[hyperindex,breaklinks]{hyperref}
\usepackage[hyperref]{emnlp-ijcnlp-2019}
\usepackage{times}
\usepackage{latexsym}

\aclfinalcopy 

\title{Paraphrasing with Large Language Models}

\author{Sam Witteveen \\
  Red Dragon AI \\
  {\tt sam@reddragon.ai} \\\And
  Martin Andrews \\
  Red Dragon AI \\
  {\tt martin@reddragon.ai} \\}

\date{}

\begin{document}
\maketitle
\begin{abstract}
Recently, large language models such as GPT-2 have shown themselves to be extremely adept at text generation and have also been able to achieve high-quality results in many downstream NLP tasks such as text classification, sentiment analysis and question answering with the aid of fine-tuning. We present a useful technique for using a large language model to perform the task of paraphrasing on a variety of texts and subjects. Our approach is demonstrated to be capable of generating paraphrases not only at a sentence level but also for longer spans of text such as paragraphs without needing to break the text into smaller chunks.
\end{abstract}

\section{Introduction}

Paraphrase generation is an NLP task that has multiple uses in content creation, question answering, translation, and data augmentation. It is a task that has been attempted for many decades using statistical and rules-based approaches \cite{McKeown1979ParaphrasingUG,Meteer1988StrategiesFE}.

We propose a system that generates paraphrased examples in an autoregressive fashion using a neural network, without the need for techniques such as top-k word selection or beam search.

We demonstrate that by using large language models we are able to produce not only paraphrases that are longer and of a higher quality than previous work, but can also paraphrase text beyond the individual sentence-level (i.e. full paragraphs at a time).

The large language models we use implement the encoder-decoder structure of the transformer architecture \cite{vaswani2017attention} which has been shown to learn different representations of language at each level of its encoding \cite{Devlin2018BERTPO}. The power of language models like GPT-2 \cite{radford2019language} and BERT allows them to develop useful representations of language which can be used far beyond just generation of the next word  \cite{DBLP:journals/corr/abs-1907-12461}. In our experiments, we have observed that the models have representations of syntax and grammar, allowing them to be fine-tuned for the task of paraphrase generation.

\section{Related Work}

Paraphrase generation has attracted a number of different NLP approaches.  These have included rule-based approaches \cite{McKeown1979ParaphrasingUG,Meteer1988StrategiesFE} and data-driven methods \cite{Madnani2010GeneratingPA}, with recently the most common approach being that the task is treated as a language translation task \cite{Bannard2005ParaphrasingWB,Barzilay2001ExtractingPF,Pang2003SyntaxbasedAO} - often performed using a bilingual corpus pivoting back and forth \cite{Madnani2010GeneratingPA,DBLP:journals/corr/PrakashHLDQLF16,mallinson2017paraphrasing}. 
Other methods proposed include more recently the use of Deep Reinforcement Learning \cite{Li2017ParaphraseGW} , supervised learning using sequence-to-sequence models \cite{Gupta2017ADG,DBLP:journals/corr/PrakashHLDQLF16} and unsupervised approaches \cite{Bowman2015GeneratingSF,Roy2019UnsupervisedPW}.

\begin{table*}[t!]
    \centering
    \begin{tabular}{|l|p{1.25\columnwidth}|l|l|l|} 
    					\hline							
     			&\multicolumn{1}{r}{} & USE & ROUGE-L & BLEU \\				
    \hline
    \textbf{\small{In:}} & A prisoner can asphyxiate himself in 90 seconds and, after eight minutes or so, he will be brain dead. & & &  \\
    \textbf{\small{Out:}} & In 90 seconds, a prisoner can asphyxiate himself and be brain dead after eight minutes or so.  & 0.9326 & 0.4706 & 0.4730\\\hline
    \small{\textbf{In:}} & The restaurant is a carved-off space up a couple of stairs to one side, dominated by faux bare-brick columns, faux-wood floors and an air of foetid despondency. & & &  \\
    \textbf{\small{Out:}} & It is a carved-off space, up a couple of flights of stairs, to the other side of the restaurant, dominated by fake bare-brick columns, fake wood floors and an air of foetid despondency. & 0.8954 & 0.5000 & 0.5348\\\hline
    \textbf{\small{In:}} & I signed a bill that made the problem worse, and I want to admit it, he said. & & &  \\
    \textbf{\small{Out:}} & He signed a bill that made the problem worse and he wants to admit it. & 0.8363 & 0.4667 & 0.5299\\\hline
    \textbf{\small{In:}} & It said the damage to the wing provided a pathway for hot gasses to penetrate the ship's thermal armor during Columbia's ill-fated reentry. & & &  \\
    \textbf{\small{Out:}} & The document says the damage to the wing provided a pathway for hot gases to penetrate Columbia's thermal armour during its fatal re-entry.  & 0.9190 & 0.4545 & 0.5445 \\
    \hline
       \end{tabular}
       \caption{Examples of GPT-2 generated paraphrased sentences with scores for each pair}
    \label{tab:examples}
\end{table*}

\section{Methodology}
In this section, we define the task, present our implementation, describe our evaluation protocol and the paraphrase datasets used.

Paraphrase generation can be described as generating a phrase which is semantically as close as possible to the original while being rewritten in new words and phrases.

\subsection{Large Language Model training}
Training language models has become a popular way of creating models suited for Transfer Learning in the field of NLP. \cite{howard-ruder-2018-universal,Peters:2018,Radford2018ImprovingLU,	DBLP:journals/corr/abs-1905-03197}. While these models are initially trained in a semi-supervised manner to predict the next word or words in a sequence, they can be fine-tuned and used for a variety of downstream NLP tasks such as text classification, sentiment analysis, tagging, and entity extraction.

More recently, large language models using transformer architectures are achieving state of the art results for many of these tasks while using less supervised data than previously needed.

One example of these large language models that has proven to be very good at text generation is GPT-2. It makes use of a transformer architecture and comes in various sizes up to 1.5 billion parameters. In these experiments, we have taken a pre-trained version of the GPT-2 model trained in a semi-supervised fashion on the WebText dataset \cite{radford2019language} of over 8 million documents with 40 GB of text in total.

\subsection{Fine-tuning for Task}

We take the GPT-2 model and fine-tune it on a supervised dataset of pre-made paraphrase examples. These examples are fed into the model as original phrase / paraphrase pairs, separated by a specific identifying sequence (such as "$>>>>$"). 

This training is done for a small number of epochs to give the model just enough examples of what the task is asking from the model : The goal being to avoid overfitting the model on the new data, while giving it sufficient exposure to the task to enable it to learn the general pattern expected. 

While we experimented with TPUs for the fine-tuning, in the end we were able to reproduce the same results on a single K-80 GPU with around 90 minutes of training.

Once the model is fine-tuned, we find that it can also produce similar paraphrase training examples if sampled from with no conditional input. To give an indication of training progress, these 'naive' paraphrases are sampled on a periodic basis during the training.

After fine-tuning on this dataset, we are then able to feed in any original phrase followed by the unique token and have the model generate paraphrases on demand.

\subsection{Candidate Generation and Selection}

After the model is trained, we then sample from the model using previously unseen sentences as conditional input. This conditional input allows us to generate multiple candidate sentences for the single original sentence.

While the quality of the paraphrases is somewhat variable, by generating multiple outputs and then scoring them, we can select just the best quality paraphrases based on a number of criteria that serve to filter our output down to a set of satisfactory results.

First, we obtain a similarity score between the generated paraphrase and the original sentence by using the Universal Sentence Encoder (USE) \cite{Cer2018UniversalSE}  to make a 512 dimensional sentence embedding for each output sentence and then compare them to the embedding of the original sentence via the cosine similarity measure.

As a second step, we measure the ROUGE-L \cite{Lin2004ROUGEAP} score of the candidate paraphrases against the original sentence and eliminate candidates with a ROUGE-L score of above 0.7 . This prevents candidates that are too close to the original sentence being chosen. After testing both cutoff scores for ROUGE-L and BLEU  \cite{papineni-etal-2002-bleu}, ROUGE-L has shown to be more useful at finding candidates that are more unique in comparison to the original sentence.

By choosing samples with sufficiently low ROUGE-L scores but as high a similarity as possible, we end up with an output that is semantically similar to the original phrase but has a unique word order when compared to the original phrase.

\subsection{Datasets}
We fine-tuned multiple versions of the model on several different datasets : 2 datasets of sentences and their matching paraphrases; and 1 dataset of paragraphs with matching paraphrases :
\begin{enumerate}

\item The MSR Paraphrase Identification dataset \cite{Dolan2004UnsupervisedCO} which consists of just over 4,000 examples of original sentences with a matching paraphrased sentence in its train set.
\item An original dataset of 10,000 sentences from online news articles along with matching paraphrases that were human-generated.
\item A further original dataset of paragraphs with corresponding paraphrased paragraphs from various entertainment, news, and food articles found online, where the paraphrases were human-generated.
\end{enumerate}

We fine-tuned 3 versions of the GPT-2 model, one corresponding to each dataset, and then made predictions using the same system outlined above.

By calculating USE, ROUGE-L and BLEU scores for each dataset we are able to quantify the quality of human-generated paraphrases and then use that as a comparison for the model’s generated sentences (see Table 2).

\begin{table}[ht]
\centering
   		 \begin{tabular}{l|llll}
        Dataset      & USE    & R-L    & BLEU   \\ \hline
        \textsc~~MSR\_train  & 0.8462 & 0.4315 & 0.4593\\
        ~~MSR\_test & 0.8415 &  0.4202 & 0.4966 \\
        ~~News dataset  & 0.8948 & 0.4686 & 0.5648 \\
        ~~Paragraphs dataset & 0.9208 & 0.4966 & 0.5762
    		\end{tabular}
    		\caption{Average USE, ROUGE-L, BLEU Scores of the datasets}
    		\label{tab:dataset}
    		\end{table} 
  		
\begin{table*}[t!]
    \centering
    \begin{tabular}{|l|p{1.5\columnwidth}|l|c|c|} 
    					\hline							
     			&\multicolumn{1}{r}{} & USE & R-L   \\				
    \hline
    \textbf{\small{In:}} & A prisoner can asphyxiate himself in 90 seconds and, after eight minutes or so, he will be brain dead. & &   \\\hline
    
    \textbf{\small{Out 1:}} & After 8 minutes, a brain fart will subdue the sufferer. & 0.524 & 0.0   \\\hline
    \textbf{\small{Out 2:}} & After 8 minutes, he will be brain-dead and his heart will stop. & 0.565 & 0.138   \\\hline
    
    \textbf{\small{Out 3:}} & A brain aneurysm can asphyxiate itself in 90 seconds and, after eight minutes, it will be dead. & 0.721 & 0.412   \\\hline
    
     \textbf{\small{Out 4:}} &  After eight minutes, a brain anesthetist can asphyxiate a prisoner in 90 seconds and for several minutes after that. & 0.758 & 0.167  \\\hline
     
  	\textbf{\small{Out 5:}} & A brain-dead prisoner canasphyxiate himself in 90 seconds and then out loud after eight minutes. & 0.809 & 0.312   \\\hline
  	
    \textbf{\small{Out 6:}} & At asphyxiation, the prisoner canasphyxiate himself in 90 seconds and, after 8 minutes, he will be brain dead. & 0.884 & 0.514  \\\hline
    
    \textbf{\small{Out 7:}} & After eight minutes, a prisoner can asphyxiate himself in 90 seconds and, after that, he will be brain dead. & 0.884 & 0.514  \\\hline
    
    \textbf{\small{Out 8*:}} & \textbf{In 90 seconds, a prisoner can asphyxiate himself and be brain dead after eight minutes or so} & \textbf{0.932} & \textbf{0.473}  \\\hline
    
    \textbf{\small{Out 9:}} & A prisoner can asphyxiate himself in 90 seconds and, after eight minutes, he will be brain dead. & 0.972 & 0.824  \\\hline
 
       \end{tabular}
       \caption{Showing Candidates Selection and Scoring - *Selected Sentence}
    \label{tab:examples_sentences}
\end{table*}

\section{Experiments}

We implemented the system described above using GPT-2 and trained it on the different datasets for various lengths of training.

To evaluate the output of the model, we randomly selected sentences from sources such as Wikipedia, news sites and entertainment sites with no matching paraphrase to use as the conditional input to the model.

\section{Results and Scoring}

When comparing our generated sentences with the average scores of the original datasets, we can see that that they compare favorably.

As discussed earlier,  we assessed the semantic similarity of the sentence meanings using Universal Sentence Encoder \cite{Cer2018UniversalSE} and compared them to the average USE score from the datasets that were trained on. This showed that the system can generate paraphrases which are semantically on par with the human-generated ones in each of the datasets.

We also compared the ROUGE-L \cite{Lin2004ROUGEAP} scores of the generated samples with the average values for the datasets which were human-generated. This again shows that our phrases are coherent and on par with human-generated paraphrases.

When we further compared the results of unfiltered examples generated by the model (Table 3) we observe that when the USE score is below 0.85 we see clear deterioration in the semantic similarity quality of the paraphrased versions. 

We also observe that if the USE score is too close to 1.0 then the ROUGE-L score also rises and the generated examples are too similar in word and phrase selection to the original sentence to be useful paraphrases.

This technique can be performed not only at sentence-level but also to generate paragraph-level paraphrases. Comparing USE and ROUGE-L scores of the generated paragraphs we see they are again on par with the human generated examples from our paragraph dataset (samples are given in the Supplemental Materials). 

Due to the pre-training of the Language Model, the model is able to generalize to and generate paraphrases for types of content it has never seen during the fine-tuning phase.

\section{Discussion}

The technique outlined in this paper shows the applicability of large language models to the paraphrasing task.  It also highlights that there is still much to be learnt about further applications of large language models, and also the approaches used to fine-tune and use them for applications. 

Most of the results from models such as GPT-2 have focused on the quality of text generation rather than quantitative methods for measuring and improving the quality of text created, to make it more consistent and usable. We propose the scoring and filtering of candidates using techniques such as we have shown with USE and ROUGE-L, may be a useful technique not just for paraphrasing but other text generation tasks.

The ability of our technique to work with long spans of text also gives it an advantage over prior work which used rule-based and other statistical approaches which performed best on shorter spans of text.

Our experiments show that  pre-training of GPT-2 on such a large amount of data in the WebText dataset allows it to 'understand' the syntax and to a degree the grammar of English allowing it to be able to quickly learn the task of paraphrasing through fine-tuning training on a small set of paraphrasing examples.

\section{Future Work}
Extrapolating from the paraphrasing results into more generalizable ideas, we hope to investigate the extent by which the representations learned in the different layers of the transformer network correspond to different parts of the linguistic hierarchy.  One possible approach to doing this would be to trace a set of 'markers' through the transformer network’s existing attention mechanism, in parallel to the text which gives rise to that structure.

In addition,  the ability of the networks to learn tasks within the span of a single context frame indicates the possibility of an inherent bias towards meta-( or one-shot) learning.  These will be the subject of further work.

\section*{Acknowledgments}

We would like to thank Google for access to the TFRC TPU program which was used in training and fine-tuning models for this paper.

\bibliographystyle{acl_natbib}
\bibliography{emnlp-ijcnlp-2019}

\cleardoublepage 
\twocolumn \begin{@twocolumnfalse}
\section*{Supplemental Materials}

\vspace{1mm}

\normalsize

\begin{table*}[ht!]
\setlength{\tabcolsep}{6.9pt}
\renewcommand{\arraystretch}{1.3}

  \centering
  
\vspace{1.5mm}
\begin{tabular}{|l|p{14.5cm}|}
	\hline
    \textbf{\small{In:}} & The first step in solving the problem is to prop the door open with tape. This means your cat will now be able to see through to the other side; your cat will likely begin using the kitty door immediately. Once your cat has gotten used to using the kitty door, remove the tape. Sometimes cats will continue to use the kitty door without any more prompting. \\
    \textbf{\small{Out:}} & The first step in resolving the problem is to prop the gate with tape. This implies that your cat will shortly be enabled to see through to the other side; your cat will likely immediately begin using the kitty door. Remove the tape when your cat has got used to using the kitty door. Sometimes cats will continue to use the kitty gate without any more command.\\
    \textbf{\small{Scores:}} & \textbf{USE:} 0.9835	~~~~~~~~~~~~~~~~~ \textbf{ROUGE-L:} 0.6102 ~~~~~~~~~~~~~~~~~ \textbf{BLEU:} 0.6858 \\\hline
    \small{\textbf{In:}} & I do well in school, and people think I am smart because of it. But it’s not true. In fact, three years ago I struggled in school. However, two years ago I decided to get serious about school and made a few changes. First, I decided I would become interested in whatever was being taught, regardless of what other people thought. I also decided I would work hard every day and never give up on any assignment. \\
    \textbf{\small{Out:}} & I do well in school, and individuals believe that I am intelligent because of it. But it's not true. In fact, three years ago I struggled in school. I decided I would become interested in whatever was taught, regardless of what other people thought. I also decided I would work every day and never give up on any task. \\
    \textbf{\small{Scores:}} & \textbf{USE:} 0.9380	~~~~~~~~~~~~~~~~~  \textbf{ROUGE-L:} 0.6929 ~~~~~~~~~~~~~~~~~ \textbf{BLEU:} 0.6221 \\\hline
    \textbf{\small{In:}} & There are three reasons why Canada is one of the best countries in the world. First, Canada has an excellent health care system. All Canadians have access to medical services at a reasonable price. Second, Canada has a high standard of education. Students are taught by well-trained teachers and are encouraged to continue studying at university. Finally, Canada's cities are clean and efficiently managed. Canadian cities have many parks and lots of space for people to live. As a result, Canada is a desirable place to live.\\
    \textbf{\small{Out:}} & There are three main reasons why Canada is one of the finest locations in the globe. First, Canada has an excellent health care system. All Canadians have access to medical care at a reasonable cost. Second, Canada has a high standard of education. Students are taught by well-trained teachers and are urged to continue studying at university. Finally, Canada's cities are clean and efficiently managed with many parks and lots of room for living, as a consequence, Canada is a desirable location.\\
    \textbf{\small{Scores:}} & \textbf{USE:} 0.9775 ~~~~~~~~~~~~~~~~~ \textbf{ROUGE-L:} 0.6832	~~~~~~~~~~~~~~~~~ \textbf{BLEU:} 0.7182 \\\hline
    
    \end{tabular}
    \caption{Sample paragraphs pairs}
    \centering
\end{table*}

\end{@twocolumnfalse}
\end{document}